\documentclass[letterpaper, 10 pt, journal, twoside]{IEEEtran}

\IEEEoverridecommandlockouts                              % This command is only needed if 
\usepackage{amsmath,amsfonts}
\usepackage{algorithmic}
\usepackage{algorithm}
\usepackage{array}
\usepackage[caption=false,font=normalsize,labelfont=sf,textfont=sf]{subfig}
\usepackage{textcomp}
\usepackage{stfloats}
\usepackage{url}
\usepackage{verbatim}
\usepackage{graphicx}
\usepackage{amssymb}
\usepackage{mathtools}
\usepackage{leftindex}
\usepackage{tabularray}
\usepackage{xcolor}
\usepackage{gensymb}
\usepackage{indentfirst}
\usepackage{colortbl}
\usepackage{caption}
\usepackage{multirow}

\usepackage{fancyhdr}

\usepackage{adjustbox}

\let\labelindent\relax
\usepackage{enumitem}

\definecolor{darkgray}{rgb}{0.3,0.3,0.3}

\UseTblrLibrary{booktabs}

\DeclarePairedDelimiter{\abs}{\lvert}{\rvert}

\makeatletter
\let\NAT@parse\undefined
\makeatother
\usepackage{hyperref}
\usepackage{cite}
\usepackage[capitalise]{cleveref}
\Crefname{equation}{Eq.}{Eqs.}
\Crefname{figure}{Fig.}{Figs.}
\Crefname{table}{Tab.}{Tabs.}
\Crefname{section}{Sec.}{Secs.}

\newcommand{\red}[1]{\textcolor{black}{#1}}

\newcommand{\insertfig}{\vspace{2pt}\includegraphics[width=0.85\linewidth]{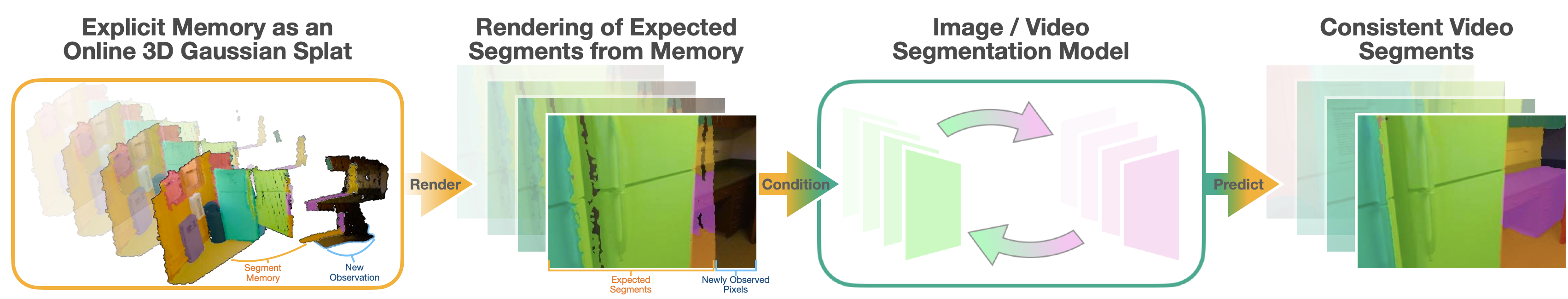}
\vspace{-4pt}
\captionof{figure}{Illustration of this paper's core insight: Using explicit memory in the form of an online 3D Gaussian splat to condition image and video segmentation models for improved video segmentation consistency.}\label{fig:teaser}
\vspace{-12pt}
}
\makeatletter
\apptocmd{\@maketitle}{\setcounter{figure}{0}\centering\insertfig}{}{}% insert the figure after authors
\makeatother

\title{
Explicit Memory through Online 3D Gaussian Splatting Improves Class-Agnostic\\Video Segmentation
}

\author{Anthony Opipari$^{1}$, Aravindhan K Krishnan$^{2}$, Shreekant Gayaka$^{2}$, Min Sun$^{2}$\\Cheng-Hao Kuo$^{2}$, Arnie Sen$^{2}$, Odest Chadwicke Jenkins$^{1}$ % <-this % stops a space
\thanks{Manuscript received: May, 15, 2025; Revised August, 13, 2025; Accepted September, 20, 2025.}%Use only for final RAL version
\thanks{This paper was recommended for publication by Editor Markus Vincze upon evaluation of the Associate Editor and Reviewers’ comments.} %
\thanks{$^{1}$University of Michigan, {\tt\small \{topipari, ocj\}@umich.edu}}%
\thanks{$^{2}$Amazon Inc., {\tt\small \{krsar, sgayaka, minnsun, chkuo, senarnie\}@amazon.com}}%
\thanks{Digital Object Identifier (DOI): see top of this page.}%
}

\begin{document}

\maketitle

\begin{abstract}
Remembering where object segments were predicted in the past is useful for improving the accuracy and consistency of class-agnostic video segmentation algorithms.
Existing video segmentation algorithms typically use either no object-level memory (e.g. FastSAM) or they use implicit memories in the form of recurrent neural network features (e.g. SAM2).
In this paper, we augment both types of segmentation models using an explicit 3D memory and show that the resulting models have more accurate and consistent predictions.
For this, we develop an online 3D Gaussian Splatting (3DGS) technique to store predicted object-level segments generated throughout the duration of a video.
Based on this 3DGS representation, a set of fusion techniques are developed, named FastSAM-Splat and SAM2-Splat, that use the explicit 3DGS memory to improve their respective foundation models' predictions.
Ablation experiments are used to validate the proposed techniques' design and hyperparameter settings.
Results from both real-world and simulated benchmarking experiments show that models which use explicit 3D memories result in more accurate and consistent predictions than those which use no memory or only implicit neural network memories.\\
Project Page: \href{https://topipari.com/projects/FastSAM-Splat/}{https://topipari.com/projects/FastSAM-Splat}

\end{abstract}


\vspace{-4pt}
\section{Introduction}
\label{sec:intro}
\vspace{-2pt}

\IEEEPARstart{O}{bject}-level semantic mapping is a crucial challenge in deploying autonomous robots in domestic human environments\red{~\cite{RUSU2008927}}.
In order to construct useful maps in these settings, robots must detect and segment objects from any category, including those that were unknown before deployment.
This challenge is compounded by the substantial variability of real-world environments, where occlusion, low illumination, duplicated, and dynamic objects are common\red{~\cite{lai3dsegmenting}}.
Even the robot's own embodiment—its camera placement, sensor modality, motion profile, etc.—is a source of variability that limits the generalization of state-of-the-art segmentation algorithms in the open-world~\cite{opipari2024mvpd}.
To address these challenges, recent research has focused on the potential for class-agnostic vision foundation models, such as SAM~\cite{kirillov2023segany} and FastSAM~\cite{zhao2023fast} to contribute to semantic mapping.
While foundation models like these have shown excellent performance in class-agnostic image segmentation, their predictions are not temporally consistent~\cite{ravi2024sam2}.
\red{Meanwhile, advances in mapping systems like Clio~\cite{maggio2024clio} have identified the consistency of image-level segments as a key enabler of open-world semantic maps.}
This observation motivates the use of \emph{video segmentation} in place of static image segmentation algorithms for \red{robotic mapping, where the consistency of object segments is critical~\cite{YANG2024105235}}.
An expanding body of work has proposed techniques to ensure these models predict consistent segments across space~\cite{cen2023segment,ying2024omniseg3d, lyu2024gaga, xu2024esam}.
Building on this emerging area of research, this paper aims to explore the following question: \textbf{Can estimated image-level segments be combined efficiently over time to enhance the consistency of downstream video segments (\cref{fig:teaser})?}

\red{In particular, we set out to improve the temporal consistency and accuracy of state-of-the-art class-agnostic video segmentation.}
Unlike traditional segmentation approaches that are trained to segment objects from a predefined closed-set of categories, class-agnostic methods enable robots to track objects from categories not seen during training.
The recent SAM2~\cite{ravi2024sam2} video segmentation model achieved state-of-the-art performance on this task without using an explicit 3D memory of its past predictions.
We hypothesize that models like SAM2 would benefit from access to a dense 3D memory of past predictions (i.e. a robot's object map).
The present study sets out to evaluate this hypothesis by developing class-agnostic video segmentation models which use explicit 3D memories to inform their predictions.

This paper proposes an approach that uses a robot's estimated semantic map to improve foundation segmentation model predictions in video segmentation.
Specifically we develop two models, referred to as \textbf{FastSAM-Splat} and \textbf{SAM2-Splat}, which build upon the FastSAM and SAM2 foundation models respectively, by integrating explicit spatial memory in the form of a 3DGS.
The proposed approach is inspired by recent online 3DGS techniques~\cite{keetha2024splatam,Matsuk2024gsslam,li2025streamgs}, which construct high-fidelity, explicit maps of a robot's environment in real-time using videos as input.
Our proposed model builds a 3DGS memory alongside its video segmentation, allowing the 3DGS to store a spatial history of past predictions.
At each step in the process, the 3DGS map can then be used to refine the foundation model's prediction either with a segment-matching algorithm (FastSAM-Splat) or a re-prompting strategy (SAM2-Splat).

This paper makes the following contributions:
\begin{enumerate}
    \item \textbf{FastSAM-Splat:} An extension of FastSAM—which originally lacks temporal memory—that integrates a 3DGS memory to improve segmentation consistency.
    \item \textbf{SAM2-Splat:} A 3DGS-based re-prompting strategy to improve SAM2—which originally uses only implicit memories—by incorporating explicit 3D memories to reduce inconsistent predictions.
    \item \textbf{Experimental} results showing models which use explicit 3D memories produce more consistent video segments than those with no or only implicit memories.
\end{enumerate}

\vspace{-4pt}
\section{Related Work}
\label{sec:relatedwork}

\textbf{Video segmentation} tasks typically fall into one of two types: semantic or class-agnostic segmentation.
In video semantic segmentation, algorithms predict pixel-level segments for each object in a video along with a classification of which categories the objects belong while video instance segmentation requires models distinguish between multiple objects of the same category~\cite{yang2019vis}. 
Many approaches for semantic video segmentation specialize to specific categories of interest~\cite{vertens2017smsnet,li2022videoknet,miao2022vipseg,yang2022tevit,xu2023viposeg,li2023tube}.
For instance, Video K-Net~\cite{li2022videoknet} learns a specialized kernel for each class and each instance.
Similarly, PAOT~\cite{xu2023viposeg} and Tube-Link~\cite{li2023tube} are transformer-based architectures that associate segments throughout the video based on which category the segments belong to.
\textbf{Class-agnostic video segmentation} tasks on the other hand, require that objects be segmented and tracked \textit{regardless of their semantic class}~\cite{siamv2021cas}.
When only a few specific objects are of interest, video object segmentation is considered~\cite{ren2007tracking,tokmakov2017vos,perazzi2017vos,caelles2017vos,cheng2017segflow}, whereas class-agnostic video instance segmentation sets out to segment and track every object of every class~\cite{opipari2024mvpd}.
Within class-agnostic video segmentation, Siam et al. proposed using motion cues in the form of optical flow to separate object segments~\cite{siamv2021cas}.
The present paper builds on these ideas to consider whether motion and depth cues can improve video segmentation in open-world robotic settings.

\textbf{Open-world segmentation} tasks consider the scenario in which robots must segment categories of objects that were unknown during training.
Many approaches have been proposed for open-world image segmentation~\cite{bucher2019zssem,danielczuk2019segmenting,zheng2021zsis,xiang2021learning,kirillov2023segany,ornek2023super,zhao2023fast} and interest in open-world video segmentation has been growing~\cite{wang2023seggpt,cheng2017segflow,du2021unseenvidseg,siamv2021cas,xu2023viposeg,cheng2023tracking,opipari2024mvpd,ravi2024sam2,ding2024sam2long}.
For distinguishing foreground and background objects in zero-shot settings, object motion through optical flow has been used~\cite{cheng2017segflow,du2021unseenvidseg,siamv2021cas}.
Wang et al.~\cite{wang2023towards,wang2024ov} and Li et al.~\cite{li2024omg} propose using language features in open-world video segmentation.
SegGPT~\cite{wang2023seggpt} proposed using an image-level foundation model and in-context coloring for open-world video segmentation.
More recently, Ravi et al.~\cite{ravi2024sam2} extended the image-level Segment Anything Model (SAM~\cite{kirillov2023segany}) with recurrent neural network features to improve video segmentation efficiency and consistency.
Similarly, GLEE~\cite{wu2024general} proposed using bipartite feature matching to extend an image-level foundation model to video segmentation tasks.
We are inspired by the open-world generalization of these foundation segmentation models but observe that they do not use explicit memory to reduce segment inconsistencies.
Ding et al.~\cite{ding2024sam2long} observed that SAM2 suffers from ``error accumulation" on long videos, particularly when objects are occluded and re-appear, and proposed a temporal memory tree for video object segmentation.
In contrast, we set out to segment and track \textit{every object} and focus on robotic use cases, where embodiment (i.e. depth and camera pose) may be used to form memories.
For these settings, FastSPAM~\cite{opipari2024mvpd} was proposed to extend the image-level FastSAM~\cite{zhao2023fast} model to class-agnostic video segmentation by tracking a set of sparse object centroids for recursive self-prompting.
Building on these ideas, this paper considers the potential for a dense 3D memory of object segments to be used to condition segment predictions and further improve video segmentation consistency.

\red{
\textbf{3D Gaussian Splatting} is a powerful representation for dense 3D scene reconstruction and fast novel-view synthesis~\cite{kerbl3Dgaussians}.
While the original 3DGS representation was limited to reconstructing scene geometry and appearance, a growing body of work has focused on encoding image-level segments in 3DGS reconstructions~\cite{shen2024flashsplat,ye2024gaussian,dou2024cosseggaussians,lyu2024gaga,qin2024langsplat,gu2024egolifter,cen2025saga}.
Gaussian grouping~\cite{ye2024gaussian} and Gaga~\cite{lyu2024gaga} assign learnable identity embeddings to each Gaussian and decode instance segments with a scene-level neural network.
CoSSegGaussians~\cite{dou2024cosseggaussians} assigns pre-trained foundation model features to the Gaussians and trains a decoder to generate instance segments.
LangSplat~\cite{qin2024langsplat} uses pre-trained language features for each Gaussian; however, these are limited to category-level segmentation.
% In place of direct feature supervision, 
SAGA~\cite{cen2025saga} and EgoLifter~\cite{gu2024egolifter} use contrastive learning to model semantic similarity between Gaussians, while FlashSplat~\cite{shen2024flashsplat} uses linear programming to assign segment identities, assuming a fixed 3DGS reconstruction as input.
A related line of work has explored segmenting 3D scenes using neural radiance fields (NeRF) rather than 3DGS~\cite{Siddiqui_2023_CVPR,lerf2023,garfield2024,ying2024omniseg3d}.
These are generally more computationally expensive than 3DGS-based methods due to their use of per-pixel ray-marching~\cite{Matsuk2024gsslam}.
Both 3DGS and NeRF-based segmentation methods rely on global reconstruction, making them unsuitable for video segmentation, where new objects must be integrated incrementally over time.
In contrast, recent work on simultaneous localization and mapping has demonstrated how 3DGS models can be constructed incrementally from video input~\cite{keetha2024splatam,Matsuk2024gsslam,li2025streamgs}.
However, these incremental 3DGS reconstruction approaches do not encode object segments.
This paper proposes augmenting online 3DGS reconstruction with instance-level segments to form an explicit 3D memory which can improve the consistency of predicted video segments.
}

\begin{figure*}[t!]
  \centering
   \includegraphics[width=0.85\textwidth]{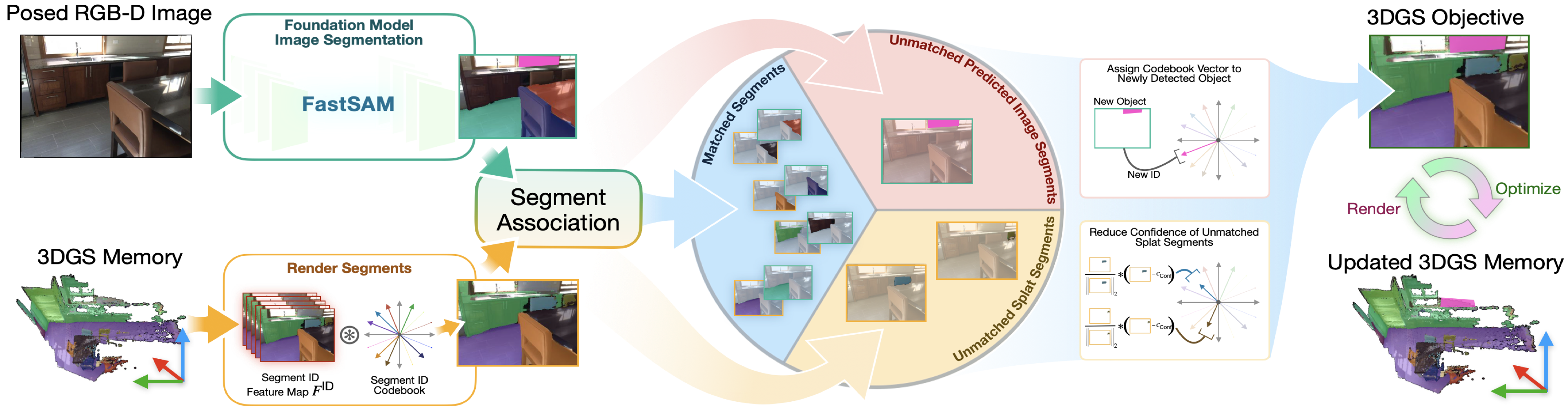}
    \vspace{-4pt}
   \caption{Illustration of the FastSAM-Splat model. Image-level `Predicted image segments' refers to an image-level segment output from the foundation image segmentation model (FastSAM).}
   \vspace{-16pt}
   \label{fig:fastsam_splat_model}
\end{figure*}

% \vspace{-4pt}
\section{Approach}
\label{sec:approach}

\vspace{-8pt}
\subsection{Preliminaries}

\textbf{Problem Definition:}
The objective of \textit{class-agnostic video instance segmentation} is to \red{detect, segment and track} every visible object instance within a video, regardless of the objects' semantic class.
Given a video sequence with $T$ image frames, $I^{1:T} = \{I^1, \ldots, I^{T}\}$, we set out to predict one segment tube for each object.
A predicted segment tube is defined as a sequence of per-frame instance masks, represented as $\hat{M}_{z_i}=\{\hat{m}^1,\ldots,\hat{m}^{T}\}_{z_i}\in \{0,1\}^{H\times W\times T}$ where $z_i$ represents a unique identifier for a single object.
Ground truth segment tubes are defined similarly, based on annotations for each frame in the video.
In addition to the input RGB images, we assume models have access to an observed depth image and camera pose for each frame.

\textbf{3DGS Representation:}
\red{We adopt a 3DGS representation to model a robot's dense 3D memories as a collection of anisotropic Gaussians.}
Each Gaussian primitive is parameterized by its 3D position ($\mu\in\mathbb{R}^3$), orientation ($q\in\mathbb{R}^4$), scale ($s\in\mathbb{R}^3$), opacity ($\sigma\in[0,1]$), and color ($c\in\mathbb{R}^3$). 
\red{In order to store semantic memories, we extend this 3DGS parameterization\cite{kerbl3Dgaussians}} by associating each Gaussian with a segment ID feature vector, $f^{\text{ID}}\in\mathbb{R}^{D_{\text{ID}}}$, which encodes the Gaussian's segment identity in a continuous feature space.
Based on this representation and given a calibrated camera pose, the 3DGS splatting process enables each Gaussian to be projected onto the camera's image space and rasterized using depth-order blending to render an output image~\cite{kerbl3Dgaussians}.
Following the same rendering process, our 3DGS can also be rendered into a segment ID feature map $F^{\text{ID}}\in\mathbb{R}^{H\times W \times D_{\text{ID}}}$.

\textbf{Online 3DGS Construction:}
\red{To use the 3DGS as a memory in video segmentation, it must be constructed incrementally as the video proceeds (online). While the original 3DGS~\cite{kerbl3Dgaussians} is constructed entirely offline, 
our approach for incremental construction is based on the Gaussian initialization and densification logic proposed in SplaTAM\cite{keetha2024splatam}, with added segment ID features for modeling semantic segments.}
At the first frame of a video, each pixel is back-projected into 3D using the depth image and camera pose, and a new Gaussian is initialized at each of these 3D points.
These Gaussians are assigned the corresponding image color, an opacity of 1, and an isotropic scale to match the size of each pixel~\cite{keetha2024splatam} given by $s=[\frac{D}{f},\frac{D}{f},\frac{D}{f}]$ where $D$ is a pixel's depth and $f$ is the camera focal length. 
The Gaussians' segment ID features are initialized based on the output of a segmentation model (e.g. FastSAM or SAM2): each detected segment is assigned a unique vector from a predefined codebook, which is \red{used} as the initial segment ID feature for any Gaussians that project into the segment mask.
In subsequent frames, new Gaussians are created for pixels that either (1) are not covered by existing Gaussians after projecting the 3DGS into the image, or (2) have a large depth disparity ($>0.15$m) between the observed and rendered 3DGS depth maps.

\textbf{Segment ID Codebook:}
Each segment is represented by a real-valued vector chosen from a predefined codebook $C=\{c_{1},\ldots,c_{N}\}\subset\mathbb{R}^{D_{\text{ID}}}$, where $N$ is the maximum number of object segments and $D_{\text{ID}}$ is the embedding space dimension.
We set out to build this codebook such that the IDs are well separated to ensure that distinct object segments are not merged during the 3DGS optimization. 
We formulate this segment ID codebook generation as an optimization problem and choose a contrastive loss as the objective function.
The chosen vectors represent segment ID using the vector direction, and model confidence using vector magnitude.
Each vector is randomly initialized then optimized with stochastic gradient descent using the following contrastive objective to maximize inter-vector distance:

\vspace{-6pt}
\begin{equation}
  L = - \min_{1 \le i \le N} \left( \min_{\substack{1 \le j \le N , \; j \ne i}} \|c_i - c_j\| \right)
\end{equation}
\vspace{-8pt}

\noindent All codebook vectors are normalized to have unit norm.

Since the 3DGS stores both RGB and segment ID features, it can be rendered into both an RGB image, $I$, as well as a segment ID feature map $F^{\text{ID}}\in\mathbb{R}^{H\times W \times D_{\text{ID}}}$.
To convert this feature map into a discrete segment map $M^{\text{ID}}\in \{1,\ldots,N \}^{H\times W}$, we compute the inner product of each pixel's feature vector with all codebook vectors and select the argument of the codebook vector which has the largest inner product and whose similarity exceeds a threshold: 

\vspace{-14pt}
\begin{align}
  m_{x,y} = \arg \max \{d(c_1,F^{\text{ID}}_{x,y}),\ldots,d( c_N, F^{\text{ID}}_{x,y})\}\\
  \text{where} \;\;\;d(c_i, F^{\text{ID}}_{x,y})=\langle c_i, F^{\text{ID}}_{x,y}\rangle * \mathbf{1}_{\langle c_i, F^{\text{ID}}_{x,y}\rangle>0.5}
\end{align}
\vspace{-14pt}

\noindent In effect, this operation applies the codebook vectors as a 2d-convolutional filter over the input feature map and filters low-confidence or low-similarity pixels.

% \vspace{-8pt}
\subsection{FastSAM-Splat}
\label{sec:fastsam_splat}
Inspired by online 3DGS algorithms, we propose an explicit memory mechanism based on the 3DGS representation to enhance segmentation consistency across frames.
To this end, we develop FastSAM-Splat, an extension of the FastSAM image segmentation model that uses an explicit 3DGS-based memory.
FastSAM-Splat incrementally builds a 3DGS memory of the scene and embeds predicted segments into its memory.
With past segments stored, the 3DGS memory is rendered into a segment feature map representing memories that can be used to condition future predictions.

\begin{figure*}[t!]
  \centering
   \includegraphics[width=0.85\textwidth]{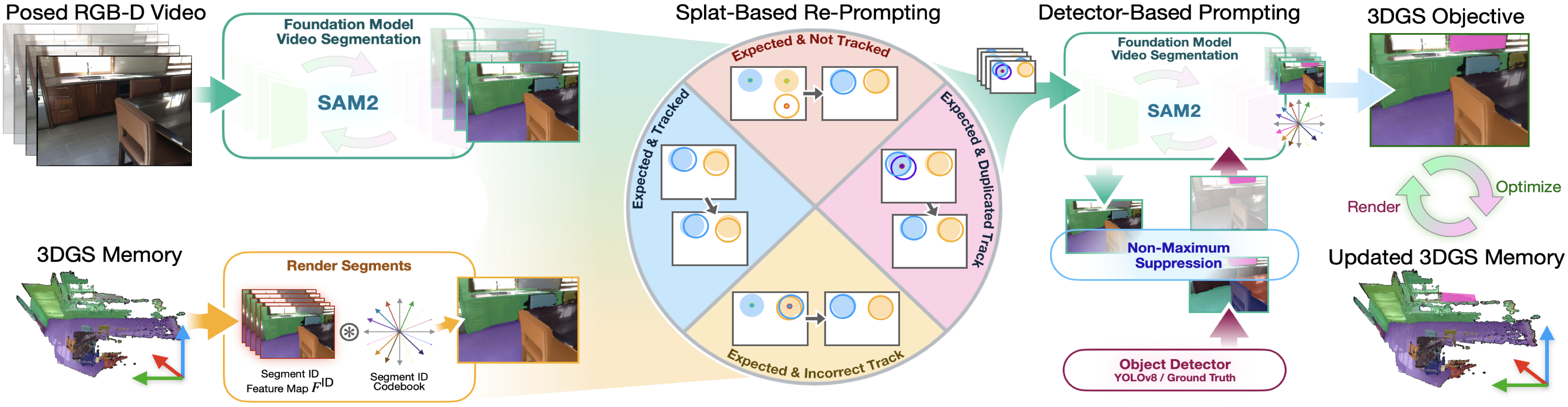}
    \vspace{-4pt}
   \caption{Illustration of the SAM2-Splat model. Splat-segment is a segment stored and rendered by Gaussian splat. Predicted image segments-output from the FastSAM/SAM2 models.}
   \vspace{-16pt}
   \label{fig:sam2_splat_model}
\end{figure*}

As illustrated in~\cref{fig:fastsam_splat_model}, FastSAM-Splat first renders the 3DGS memory into a set of 2D segments.
These rendered segments are then matched with predicted image segments from FastSAM using the Hungarian algorithm, which solves for a pairwise matching to maximize the sum total F-score across all matched segment pairs.
This association process yields three categories of segments:
\begin{itemize}
    \item Matched segments: A segment from the 3DGS which matches to a predicted image segment from FastSAM.
    \item Unmatched predicted image segments: A segment predicted by FastSAM with no associated 3DGS segment.
    \item Unmatched splat segment: A decoded segment from the 3DGS memory that has no corresponding image segment predicted by FastSAM.
\end{itemize}
Based on this segment association, FastSAM-Splat fuses the 3DGS segments with FastSAM's predicted image segments to form a final segment map used as its output prediction and as its online objective to update the 3DGS with new segment features.
The fusion process is as follows:
\begin{itemize}
    \item Matched segments inherit the existing codebook vector of their associated 3DGS segment.
    \item Unmatched predicted image segments are randomly assigned an unused codebook vector.
    \item Unmatched splat segments retain their feature ID's direction with linearly reduced magnitude according to:
\end{itemize}
\vspace{-6pt}
\begin{align}
  f^{\text{ID}'} = \frac{f^{\text{ID}}}{||f^{\text{ID}}||} * (||f^{\text{ID}}||-C_{\text{conf}})
\end{align}
\noindent where $C_\text{conf}$ is a scalar constant used to decrement the 3DGS confidence when segments are not consistently detected by FastSAM ($C_\text{conf}=0.1$ in this paper).
Once each segment is assigned a feature ID, the binary segment representations are converted into segment feature maps. 
These feature maps are aggregated into a final fused segment feature map $\hat{F}^{\text{ID}}\in\mathbb{R}^{H\times W \times D_{\text{ID}}}$.
This fused feature map is decoded using the codebook for the model's final prediction and serves as an optimization objective to update the 3DGS memory.
The 3DGS segment features are updated using stochastic gradient descent for $N_{\text{opt}}=20$ steps with the following loss:

\vspace{-14pt}
\begin{align}
  L = \lambda_{\text{mag}} \text{MSE}(||\hat{F}^{\text{ID}}||, ||F^{\text{ID}}||) +\lambda_{\text{dir}}(1-S_{C}(\hat{F}^{\text{ID}}, F^{\text{ID}}))\nonumber
\end{align}

\noindent where MSE is mean-squared error, $S_C$ denotes cosine similarity, $\hat{F}^{\text{ID}}$ is the fused objective map, and $F^{\text{ID}}$ is the rendered 3DGS feature map. Hyperparameters $\lambda_{\text{mag}}$ and $\lambda_{\text{dir}}$ control the relative contribution of magnitude and direction loss terms ($\lambda_{\text{mag}}=50.0$, $\lambda_{\text{dir}}=1.0$ in this paper).
This objective ensures that the 3DGS memory is aligned with the current frame's predictions for use at the next time step.

\vspace{-8pt}
\subsection{SAM2-Splat}
\label{sec:sam2_splat}

In addition to \red{image-based} models like FastSAM, we explore the potential for explicit 3D memory to improve video segmentation models.
To this end we extend SAM2 with a 3DGS-based memory mechanism, resulting in the SAM2-Splat model.
The core insight of SAM2-Splat is to use the 3DGS memory to identify objects that SAM2 has either failed to track or assigned an incorrect track ID (i.e. "error accumulation"~\cite{ding2024sam2long}), and to then re-prompt SAM2 using the 3DGS memory as a correction.

\begin{figure}[b]
    \vspace{-12pt}
  \centering
        \includegraphics[width=0.85\columnwidth]{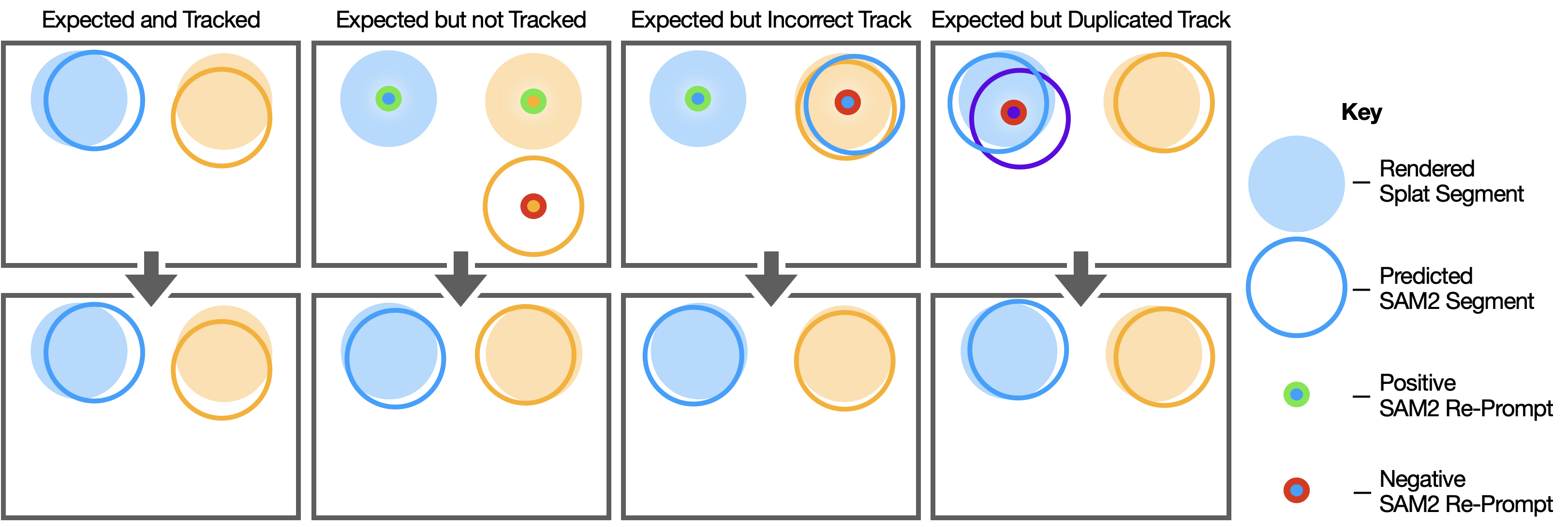}
    \vspace{-4pt}
   \caption{Illustration of the SAM2-Splat re-prompting strategy. }
    \vspace{-8pt}
    \label{fig:reprompt}
\end{figure}

SAM2 requires object prompts—as points, bounding boxes, or masks—in order to begin segmenting and tracking an object.
In this work, we use either ground truth object masks or FastSAM detections as a source of mask prompts.
On the first frame, all detected masks are used as prompts. 
On subsequent frames, non-maximum suppression is applied (\cref{sec:implementation}) to avoid tracking redundant objects.
Unlike FastSAM, SAM2 inherently predicts track IDs for each object it is tracking.
As illustrated in~\cref{fig:sam2_splat_model}, SAM2-Splat uses these predicted track IDs to associate SAM2's predictions with corresponding 3DGS segments, in-place of the Hungarian matching.
At each frame, SAM2 predicts a set of tracked segments.
SAM2-Splat compares the segments from SAM2 with those rendered from the 3DGS memory \red{as shown in~\cref{fig:reprompt}}.
This comparison sets out to identify:

\begin{enumerate}
    \item Segments that were expected but not tracked by SAM2.
    \item Segments that were tracked but with an incorrect ID.
    \item Segments that were predicted as a duplicated track.
\end{enumerate}

For each of these errors, SAM2-Splat then re-prompts SAM2 by sampling a positive and/or negative point-prompt from the relevant region of the image.
The re-prompts are designed to correct SAM2's predictions for a given object.
For example, in the third column of~\cref{fig:reprompt}, SAM2 has assigned an incorrect track ID to its predicted blue segment.
This error can be corrected by re-prompting SAM2 with a positive point sampled from the rendered 3DGS memory of the blue object and a negative point sampled from the region that SAM2 incorrectly assigned to the blue ID.
After re-prompting, SAM2 is updated with any new object detectionsand the 3DGS memory is updated using the same process described for FastSAM-Splat.
The final segment prediction for SAM2-Splat is the updated segments produced by SAM2 after it is re-prompted and incorporates new detections.

% \vspace{-4pt}
\section{Experiments}
\label{sec:exp}
\vspace{-4pt}

\textbf{Datasets:}
\label{sec:datasets}
We conduct experiments on both the ScanNet~\cite{dai2017scannet} and MVPd~\cite{opipari2024mvpd} datasets.
ScanNet includes 1,201 RGB-D videos for training and 312 videos for validation.
All videos in ScanNet are collected from real-world environments where a domestic robot would be expected to operate.
The 312 validation videos are used for testing in this paper.
We follow AnyView~\cite{wu2023anyview} and Memory-Based Adapters~\cite{xu2024memory} to downsample the videos from ScanNet into the ScanNet-MV subset.
Videos in ScanNet-MV are downsampled to a resolution of 240x320 and at 1/20th the original frame rate.

In addition to ScanNet, we use the MVPd dataset as a large-scale benchmark designed for class-agnostic video segmentation.
This dataset provides 16,200 videos for training and 1,741 videos for testing.
Each video in MVPd has RGB-D camera data at 480x640 resolution simulated from 180 scenes with ground truth camera pose and depth.
Notably, MVPd videos vary between 300 and 600 frames and contain annotations for more than 94 objects per video, on average.
Thus MVPd provides a rigorous benchmark covering a very large number of objects per video in a domestic robot setting.

\textbf{Metrics:}
\label{sec:metric}
For experiments using MVPd, the Video Segmentation Quality (VSQ) metric~\cite{opipari2024mvpd} is used to evaluate all models.
VSQ is designed for the class-agnostic video segmentation task and sets out to measure the overlap between predicted and ground truth segment tubes. 
More specifically, VSQ$^k$ measures the overlap between each ground truth and predicted segment tube of length $k$ frames as follows:

\vspace{-6pt}
\begin{equation}
  \text{VSQ}^k = \frac{\sum_{(u,\hat{u})\in TP} IoU(u,\hat{u})}{\abs{TP}+\frac{1}{2}\abs{FP}+\frac{1}{2}\abs{FN}}
  \label{eq:VSQ^k}
\end{equation}
% \vspace{-8pt}
\noindent where the Hungarian algorithm is used to match each predicted tube to at most one ground truth tube to determine the true positives (TP), false positives (FP) and false negatives (FN).
The aggregate VSQ score is the average VSQ$^k$ over a set of window sizes, $K$.
This paper uses $K=\{1,5,10,15\}$ and VSQ$^k$ is calculated using a window stride of $15$ frames.

For benchmarking experiments on ScanNet-MV, we use both VSQ and the Segmentation and Tracking Quality (STQ) metric~\cite{weber2021step}.
STQ is computed as the geometric mean between an association quality (AQ) component and a segmentation quality (SQ) component.
AQ measures the accuracy and consistency of a model's track assignment predictions while SQ measures the semantic segmentation accuracy of a models predictions.
Importantly, in the class-agnostic setting SQ reduces to the IoU between all predicted and labeled pixels.
While both VSQ and STQ set out to measure segment accuracy and consistency, only VSQ penalizes models that incorrectly predict overlapping segments while only STQ measures segmentation consistency across the entire video.

\begin{table}[b]
\vspace{-8pt}
  \centering
  \resizebox{\columnwidth}{!}{
\begin{tblr}{
        colspec={Q[l]Q[l]Q[c]Q[c]Q[c]Q[c]Q[c,m]Q[c,m]Q[c,m]Q[c,m]},
        cell{1}{1} = {r=2}{m},
        cell{1}{2} = {r=2}{m},
        cell{1}{7} = {r=2}{m},
        cell{1}{8} = {r=2}{m},
        cell{1}{9} = {r=2}{m},
        cell{1}{10} = {r=2}{m},
    }
    \toprule
     Method & Backbone & \SetCell[c=4]{c}{{{VSQ$^k$ with $k$-Frame Window}}} & & & & VSQ & STQ & AQ & SQ\\
     \cmidrule[lr]{3-6}
    & & $k=1$ & $k=5$ & $k=10$ & $k=15$ & & \\
    \midrule
    FastSAM & YOLOv8 & 46.98 & 39.19 & 34.28 & 31.21 & 37.92 & 28.17 & 19.74 & 40.19 \\
    FastSPAM & YOLOv8 & 50.56 & 42.78 & 37.33 & 34.03 & 41.18 & 30.33 & 21.86 & 42.09 \\
    SAM2 & YOLOv8 & 45.53 & 41.41 & 39.28 & 37.89 & 41.03 & 33.43 & 27.47 & 40.68 \\
    SAM2 & GT & 59.32 & 52.68 & 49.86 & 47.81 & 52.42 & 33.83 & 35.96 & 31.82 \\
    \midrule
    FastSAM-Splat & YOLOv8 & 48.07 & 44.15 & 41.98 & 40.75 & 43.74 & 38.39 & 32.83 & \textbf{44.89} \\
    SAM2-Splat & YOLOv8 & 47.70 & 44.16 & 42.14 & 41.02 & 43.76 & 35.01 & 30.68 & 39.94 \\
    SAM2-Splat & GT & \textbf{62.85} & \textbf{56.43} & \textbf{54.03} & \textbf{52.35} & \textbf{56.42} & \textbf{39.19} & \textbf{41.99} & 36.58 \\
    \bottomrule
\end{tblr}
}
\vspace{-4pt}
  \caption{Comparing FastSAM, FastSPAM, and SAM2 with the proposed FastSAM-Splat and SAM2-Splat models on the ScanNet-MV dataset for class-agnostic video instance segmentation.}
  \label{tab:exp_scannet}
  % \vspace{-12pt}
\end{table}

\textbf{Implementation Details:}
\label{sec:implementation}
FastSAM is trained on a single RTX A6000 GPU using a batch size of 16 images until convergence.
All other hyperparameters of FastSAM are left unchanged.
For all experiments using SAM2, the SAM2.1 Hiera-B+ model is used.
To avoid prompting SAM2 with redundant objects detected by YOLOv8, we use non-maximum suppression which ignores any YOLOv8 detections that have an IoU above 0.1 with any of SAM2's segments.

For experiments on ScanNet-MV, we crop a 5-pixel border out of each image due to missing RGB values along the image boundary that result from aligning the RGB and depth images.
The resulting input resolution of images on ScanNet-MV is 230x310. For experiments on MVPd, the FastSAM-Splat and SAM2-Splat models use resized images at 240x320 resolution to reduce their memory consumption.

\vspace{-4pt}
\subsection{Real-world Benchmark}
\label{sec:exp_scannet}

This experiment is designed to understand the effectiveness of the proposed models on a real-world dataset in which depth and camera pose represent the distribution a robot will expect in deployment.
More specifically, we set out to test two hypotheses: 1) using explicit 3D memory will improve the consistency of both image and video segmentation models, and 2) the improvement will be more pronounced in the image segmentation model—which does not use any form of temporal memory—than the video segmentation model, which uses implicit neural network memories.
For this test, we use the ScanNet-MV dataset, FastSAM as the baseline image segmentation model, and two variants of SAM2 as the baseline video segmentation model. 
One version of SAM2 uses YOLOv8 as a detector with non-maximum suppression (\cref{sec:implementation}) and another version prompts SAM2 for each object on the frame they become visible using ground truth segments, following the original usage~\cite{ravi2024sam2}.

\begin{figure}[t]

  \centering
    % \begin{subfigure}[t]{\textwidth}
        % \centering
        \includegraphics[width=0.95\columnwidth]{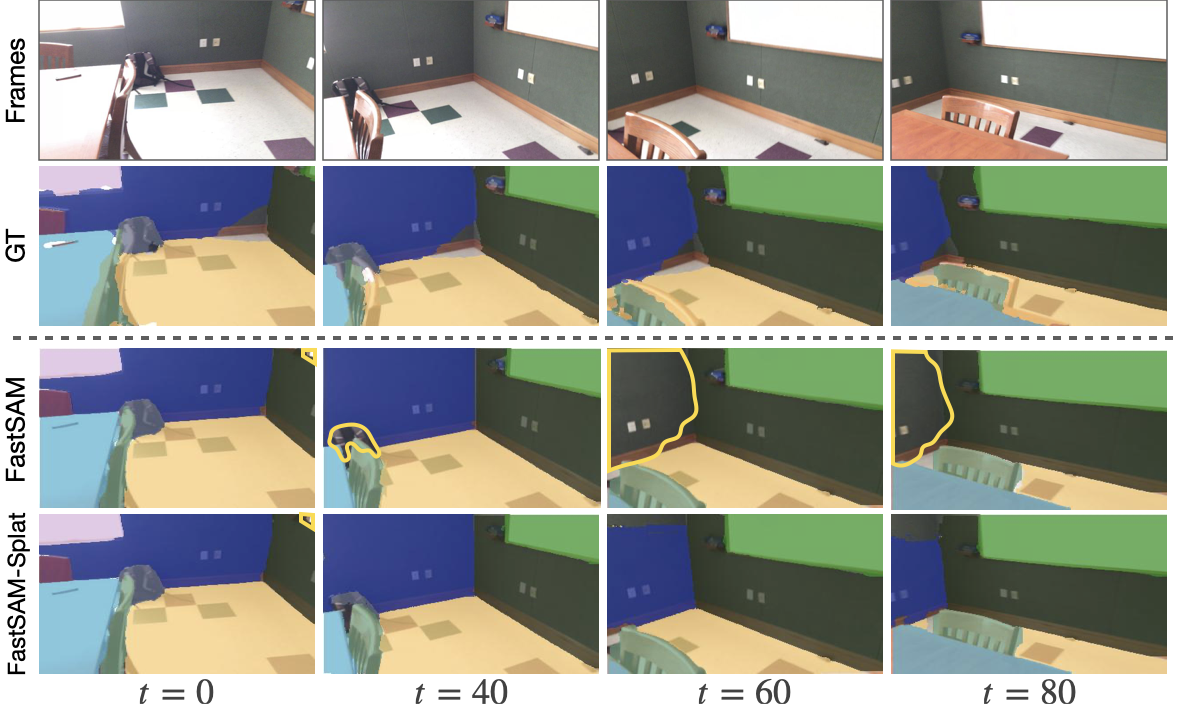}
    \vspace{-4pt}
   \caption{Qualitative comparison of the FastSAM and FastSAM-Splat models showing FastSAM-Splat has fewer segment inconsistencies. False negative inconsistencies highlighted in yellow outline.}
    \vspace{-14pt}
    \label{fig:qual_fastsamsplat}
\end{figure}

Results from this experiment are in~\cref{tab:exp_scannet} and~\cref{fig:qual_fastsamsplat}, showing FastSAM-Splat, which uses a dense 3DGS memory, achieves an improvement of +5.82\% VSQ and +10.22\% STQ over the FastSAM baseline, which does not use any temporal memories.
Moreover, the dense 3D memory used by the SAM2-Splat model leads to an improvement of +2.73\% VSQ and +1.58\% STQ over SAM2 which uses YOLOv8 detections and +4.00\% VSQ and +5.36\% STQ over the SAM2 baseline which uses ground truth detections.
\textbf{These results confirm our first hypothesis that image and video segmentation models, which both lack an explicit memory representation, result in improved segmentation consistency once given access to an explicit 3DGS memory.}

\begin{figure}[t]

  \centering
    % \begin{subfigure}[t]{\textwidth}
        % \centering
        \includegraphics[width=0.95\columnwidth]{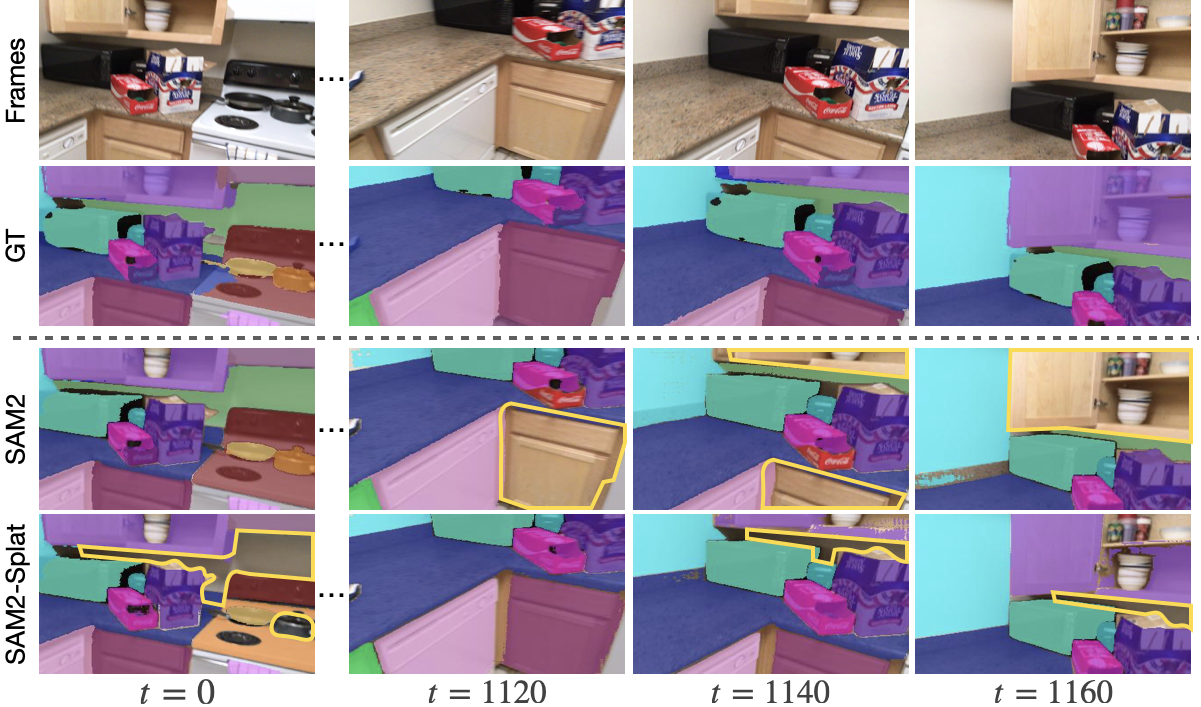}
    \vspace{-4pt}
   \caption{Comparison of SAM2 and SAM2-Splat on a sequence from ScanNet-MV showing the re-prompting mechanism reduces false negative flickering. False negatives highlighted in yellow outline.}
    \vspace{-10pt}
    \label{fig:qual_sam2splat}
\end{figure}

Next, we tested the second hypothesis that including the explicit 3D memory would result in a more substantial improvement for the image segmentation baseline than the video segmentation baseline. 
This hypothesis stems from our observation that FastSAM uses no memories while SAM2 uses recurrent neural network features as an implicit memory.
Comparing the improvement in VSQ and STQ that results from providing the 3DGS memory to FastSAM with the improvement that results from providing the memory to SAM2 shows that FastSAM-Splat led to larger improvements, even when controlling for the source of detections used by SAM2.
In the case of YOLOv8 detections, FastSAM-Splat led to a larger improvement of +3.09\% VSQ and +8.64\%STQ, while in the case of ground truth detections FastSAM-Splat again led to a larger improvement by +1.82\% VSQ and +4.86\% STQ (\cref{fig:qual_sam2splat}).
\textbf{These results confirm our second hypothesis and suggest that incorporating an explicit 3D memory during the training process for both image and video segmentation models could lead to further improvements in segmentation consistency and accuracy.}

\begin{table}[b]
\vspace{-8pt}
  \centering
  \resizebox{\columnwidth}{!}{
\begin{tblr}{
        colspec={Q[l]Q[c,m]Q[c,m]Q[c,m]Q[c,m]},
    }
    \toprule
     Method & $D_{\text{ID}}=1$ & $D_{\text{ID}}=4$ & $D_{\text{ID}}=7$ & $D_{\text{ID}}=14$ & $D_{\text{ID}}=28$\\
    \midrule
    FastSAM-Splat & 20.00 / 15.47 & 42.29 / 36.97 & 43.08 / 37.93 & 43.66 / 38.38 & \textbf{43.74} / \textbf{38.39}\\
    \bottomrule
\end{tblr}
}
\vspace{-4pt}
  \caption{Segment ID feature ablation experiment to evaluate the impact of feature vector dimension on VSQ/STQ measures of video segmentation performance. Vectors of dimension 1 ($D_{\text{ID}}=1$) use a single integer to represent each segment while vectors of encreasing dimension ($D_{\text{ID}}\in\{4,7,15,28\}$) use real valued vectors.}
  \label{tab:exp_features}
  % \vspace{-12pt}
\end{table}

\vspace{-4pt}
\subsection{Segment ID Feature Ablation}
\label{sec:exp_features}

This experiment is designed to study the impact of the segment ID features used by FastSAM-Splat.
In particular, we are interested in how the type (i.e. integer or floating point) and dimensionality of the segment ID feature vectors impacts downstream video segmentation accuracy and consistency.
Results from this ablation are shown in~\cref{tab:exp_features}.
These results show that using a single integer to represent each segment identity ($D_{\text{ID}}=1$) results in the lowest VSQ and STQ performance of 20.00\% and 15.47\% respectively.
Moving to a real-valued vector representation ($D_{\text{ID}}=4$) that is optimized using a contrastive loss leads to a substantial improvement of +22.29\% VSQ and +21.50\% STQ.
Further increasing the dimensionality of the segment ID features correlates with improved VSQ and STQ performance, however the performance begins to saturate beyond $D_{\text{ID}}=14$.
For example, increasing the dimension from $D_{\text{ID}}=4$ to $D_{\text{ID}}=14$ results in an improvement of +1.37\% VSQ and +1.41\% STQ while moving from $D_{\text{ID}}=14$ to $D_{\text{ID}}=28$ results in only +0.08\% VSQ and +0.01\% STQ.
\textbf{This shows that the optimized codebook representation outperforms a naive integer based representation for segment IDs and that increasing the dimensionality of segment ID features results in improved video segmentation consistency but saturation occurs beyond $D_{\text{ID}}=14$.}

\begin{table}[b]
\vspace{-8pt}
  \centering
  \resizebox{0.95\columnwidth}{!}{
\begin{tblr}{
        colspec={Q[l]Q[c,m]Q[c,m]Q[c,m]Q[c,m]},
    }
    \toprule
     Method & Not Tracked & Incorrect Track & Duplicated Track & VSQ / STQ\\
    \midrule
    SAM2 & & & & 41.03 / 33.43 \\
    \midrule
    SAM2-Splat & \checkmark & & & 41.09 / 34.19\\
    SAM2-Splat & & \checkmark & & 41.73 / 33.15 \\
    SAM2-Splat & & & \checkmark & 43.05 / 33.78 \\
    SAM2-Splat & \checkmark & \checkmark & \checkmark & \textbf{43.76} / \textbf{35.01} \\
    \bottomrule
\end{tblr}
}
\vspace{-4pt}
  \caption{Ablation experiment to understand the effectiveness of each re-prompt category individually and together within the SAM2-Splat (YOLOv8) model. Models are evaluated on the class-agnostic video instance segmentation task on the ScanNet-MV validation set with VSQ / STQ metrics.}
  \label{tab:exp_reprompt_category}
  \vspace{-12pt}
\end{table}

\vspace{-4pt}
\subsection{SAM2-Splat Re-Prompting Ablation}

Next, we performed a set of ablation experiments to validate the re-prompting strategy used by SAM2-Splat and to understand the impact of key re-prompting hyperparameters on video segmentation accuracy consistency.
The first hyperparameter studied is the logic used to formulate the re-prompting clicks.
The SAM2-Splat model generates re-prompting clicks based on inconsistencies identified between the 3DGS memory and the predictions made by the underlying SAM2 model.
These inconsistencies fall into three categories (\cref{sec:sam2_splat}): 1) Segments that were expected in the 3DGS memory but not tracked by SAM2 (Not Tracked), 2) Segments that were expected by the 3DGS and tracked by SAM2 but with an incorrect ID (Incorrect Track), and 3) Segments that were expected by the 3DGS memory but predicted as multiple duplicated tracks by SAM2 (Duplicated Track).
In this ablation, SAM2-Splat was evaluated on the validation set of ScanNet-MV under varying re-prompting configurations.
In each configuration, only a subset of the categories were used to formulate re-prompts and the resulting variants were compared to the default model, which uses all three categories of re-prompts, as well as the baseline SAM2, which uses no re-prompts.
The results are included in~\cref{tab:exp_reprompt_category}, showing that using only one category of re-prompt results in modest changes to the VSQ and STQ performance of SAM2-Splat when compared to the baseline SAM2 model.
Notably, the results show that using all three categories of re-prompts leads to the highest performing SAM2-Splat model in terms of both VSQ and STQ with 43.76\% and 35.01\% respectively.
\textbf{These results show that each category of SAM2-Splat's re-prompts improves SAM2 inconsistencies, and that combining all re-prompting categories leads to the largest improvement in segmentation consistency.}

\begin{table}[t]
% \vspace{-8pt}
  \centering
  \resizebox{0.9\columnwidth}{!}{
\begin{tblr}{
        colspec={Q[l]Q[c,m]Q[c,m]Q[c,m]Q[c,m]},
    }
    \toprule
     Method & Backbone & 1-click & 3-click & 5-click \\
    \midrule
    SAM2-Splat & YOLOv8 &  42.95 / 34.03 & \textbf{43.86} / 34.99 & 43.76 / \textbf{35.01} \\
    SAM2-Splat & GT & 54.89 / 37.29 & 56.10 / 38.72 & \textbf{56.42} / \textbf{39.19} \\
    \bottomrule
\end{tblr}
}
\vspace{-4pt}
  \caption{Ablation experiment on click-based re-prompting of the SAM2-Splat model. Models are evaluated on the ScanNet-MV validation set with VSQ / STQ reported for each setting.}
  \label{tab:exp_reprompt_number}
  \vspace{-12pt}
\end{table}

Another important hyperparameter when re-prompting SAM2 is the number of re-prompts used.
To understand how this impacts video segmentation consistency, we evaluated SAM2-Splat on ScanNet-MV using a varying number of re-prompts per object and report the results in~\cref{tab:exp_reprompt_number}.
These results show that for both SAM2-Splat models, using more than 1 re-prompt leads to better segmentation consistency as measured by VSQ and STQ.
For example, the SAM2-Splat model which uses ground truth detections improved by 1.21/1.43\% VSQ/STQ when moving from 1 to 3 re-prompts and by 0.32/0.47\% when moving from 3 to 5 re-prompts.
\textbf{These results indicate that using more re-prompts is generally better than fewer but improvement saturates when more than 3 re-prompts per object are used.}

\vspace{-4pt}
\subsection{Simulated Benchmark}
\label{sec:exp_mvpd}

This experiment sets out to establish the effectiveness FastSAM-Splat on a large-scale video instance segmentation dataset with additional baseline models.
To address this question, we evaluated FastSAM-Splat on the MVPd benchmark.
Quantitative results are included in~\cref{tab:exp_mvpd}, showing FastSAM-Splat achieves the highest VSQ score of 56.76\% which is +4.86\% higher than the next-best model.
Due to the high memory-consumption of SAM2 and the large number of objects in MVPd videos (up to 281), we were unable to benchmark SAM2 and SAM2-Splat models on this dataset.
\textbf{These benchmarking results provide additional evidence that the dense 3D memory of FastSAM-Splat leads to enhanced video segmentation accuracy and consistency.}

\begin{table}[t]
% \vspace{-2pt}
  \centering
  \resizebox{0.95\columnwidth}{!}{
\begin{tblr}{
        colspec={Q[l]Q[c]Q[c]Q[c]Q[c]Q[c]Q[c,m]},
        cell{1}{1} = {r=2}{m},
        cell{1}{2} = {r=2}{m},
        cell{1}{7} = {r=2}{m}
    }
    \toprule
     Method & Backbone &\SetCell[c=4]{c}{{{VSQ$^k$ with $k$-Frame Window}}} & & & & VSQ\\
     \cmidrule[lr]{3-6}
    & & $k=1$ & $k=5$ & $k=10$ & $k=15$ & \\
    \midrule
    Video K-Net & ResNet50 & 49.65 & 49.60 & 38.10 & 29.56 & 41.73 \\
    Video K-Net & Swin-base & 50.83 & 50.50 & 38.71 & 30.05 & 42.52 \\
    Tube-Link & ResNet50 & 45.13 & 20.50 & 16.08 & 14.05 & 23.94 \\
    Tube-Link & Swin-large & 48.17 & 21.14 & 16.60 & 14.55 & 25.12 \\
    OV2Seg & ResNet50 & 39.96 & 38.52 & 37.51 & 36.71 & 38.18 \\
    OV2Seg & Swin-base & 40.22 & 38.58 & 37.43 & 36.52 & 38.19 \\
    FastSAM & YOLOv8 & \textbf{61.18} & 52.03 & 44.38 & 39.09 & 49.17 \\
    FastSPAM & YOLOv8 & 60.68 & 54.10 & 48.48 & 44.35 & 51.90 \\

    \midrule
    FastSAM-Splat & YOLOv8 & 60.91 & \textbf{57.68} & \textbf{55.20} & \textbf{53.26} & \textbf{56.76} \\
    \bottomrule
\end{tblr}
}
\vspace{-4pt}
  \caption{Evaluating FastSPAM-Splat on the class-agnostic video instance segmentation benchmark of the MVPd dataset. Models are evaluated on the test videos held-out from training.}
  \label{tab:exp_mvpd}
  \vspace{-12pt}
\end{table}

\vspace{-4pt}
\subsection{Efficiency Benchmark}
\label{sec:exp_efficiency}

We evaluate the runtime and memory efficiency of each baseline model and report the results in~\cref{tab:efficiency}.
These results show that FastSAM-Splat runs at a 2.84 FPS frame rate, which is comparable to SAM2's frame rate of 2.83 FPS but slower than each of the alternative baseline models.
The SAM2-Splat model was measured to run at 1.46 FPS however we hypothesize this could be improved if the prompting API of SAM2 allowed for prompting multiple objects in-parallel.
\red{These results show that improving segmentation consistency with a 3DGS memory is possible but comes with a tradeoff in terms of higher processing latency.}
One limitation of both FastSAM-Splat and SAM2-Splat is their increased memory consumption—FastSAM-Splat used 28M parameters on average to store the explicit 3DGS memory for videos on the ScanNet-MV benchmark while SAM2-Splat used 26M parameters.
\textbf{While the memory consumption can be improved with only minor decreases in video segmentation consistency by using feature ID vectors with fewer dimensions(\cref{sec:exp_features}), these results suggest that future work to optimize 3DGS storage would be beneficial for video segmentation appraoches that seek to use 3DGS as a memory mechanism.}

\begin{table*}[!h]
  \centering
  \resizebox{0.9\textwidth}{!}{
\begin{tblr}{
        colspec={Q[l]Q[c]Q[c]Q[c]Q[c]Q[c]Q[c]Q[c]Q[c]Q[c]Q[c]Q[c]},
    }
    \toprule
     Model & \SetCell[c=2]{c}{{{Video K-Net}}} & & \SetCell[c=2]{c}{{{Tube-Link}}} &  & \SetCell[c=2]{c}{{{OV2Seg}}} & & FastSAM & FastSPAM & SAM2 & FastSAM-Splat & SAM2-Splat \\
     \cmidrule[lr]{2-3}
     \cmidrule[lr]{4-5}
     \cmidrule[lr]{6-7}
     \cmidrule[lr]{8}
     \cmidrule[lr]{9}
     \cmidrule[lr]{10}
     \cmidrule[lr]{11}
     \cmidrule[lr]{12}
    Backbone & ResNet50 & Swin-base & ResNet50 & Swin-large & ResNet50 & Swin-base & YOLOv8 & YOLOv8 & Hiera-B+ & YOLOv8 &  Hiera-B+ \\
    \midrule
    FPS & 40.54 & 26.17 & 4.88 & 4.39 & 15.80 & 13.28 & 64.57 & 6.60 & 2.83 & 2.84 & 1.46 \\
    Params (M) & 26 & 88 & 26 & 197 & 26 & 88 & 72 & 72 & 81 & 72+28 & 81+26 \\
    \bottomrule
\end{tblr}
}
  \caption{Comparing models' runtime in frames per second (FPS) and complexity in parameter count (M). For FastSAM-Splat and SAM2-Splat, the parameter count is broken into backbone parameter count + 3DGS parameter count.}
  \vspace{-8pt}
  \label{tab:efficiency}
\end{table*}

% \vspace{-4pt} 
\section{Conclusion}
% \vspace{-4pt}
This paper provides three core contributions: (1) a FastSAM-Splat model integrating 3DGS memories with FastSAM, (2) a SAM2-Splat model and re-prompting strategy to improve SAM2 video segmentation performance, and (3) empirical benchmarking experiments validating the performance of each model on simulated and real-world benchmarks.
The experiments demonstrate that incorporating explicit memories stored in a 3DGS enables an image-based model like FastSAM to outperform state-of-the-art video segmentation models. Furthermore, the explicit memory mechanism is also shown to be useful for improving the consistency of video segmentation models that do not use explicit 3D memories.
Based on these results, we envision future work to address remaining limitations of the proposed models, such as its reliance on input pose and depth observations, the lack of global optimization for the 3DGS memory, \red{and implementing 3DGS optimizations for improved runtime}.

\vspace{-2pt}

\bibliographystyle{IEEEtran}
\bibliography{main}

\end{document}